\documentclass[runningheads]{llncs}

\usepackage[final,year=2024,ID=7362]{eccv}

\usepackage{eccvabbrv}

\usepackage{graphicx}
\usepackage{booktabs}
\usepackage{algpseudocode}
\usepackage{algorithm}
\usepackage{multirow}
\usepackage{cite}
\usepackage{subcaption}
\usepackage[accsupp]{axessibility}  

\usepackage[pagebackref,breaklinks,colorlinks,citecolor=eccvblue]{hyperref}

\usepackage{orcidlink}
\usepackage{xspace}
\newcommand{\dataset}{DataVisualWorkflow\xspace}
\newcommand{\framework}{UIMTCon\xspace}
\newcommand{\nframework}{UISTCon\xspace}

\def\etal{\emph{et al}\onedot}

\begin{document}

\title{Computer User Interface Understanding. A New Dataset and a Learning Framework} 

\titlerunning{Computer User Interface Understanding}

\author{Andrés Muñoz\inst{1}\and
Daniel Borrajo\inst{1}}

\authorrunning{Andrés Muñoz et al.}


\institute{J.P. Morgan AI Research\\
\email{andres.munozgarza@jpmorgan.com}\\
\email{daniel.borrajo@jpmchase.com}}

\maketitle

\begin{abstract}
  User Interface (UI) understanding has been an increasingly popular topic over the last few years. So far, there has been a vast focus solely on web and mobile applications. In this paper, we introduce the harder task of computer UI understanding. With the goal of enabling research in this field, we have generated a dataset with a set of videos where a user is performing a sequence of actions and each image shows the desktop contents at that time point. We also present a framework that is composed of a synthetic sample generation pipeline to augment the dataset with relevant characteristics, and a contrastive learning method to classify images in the videos. We take advantage of the natural conditional, tree-like, relationship of the images' characteristics to regularize the learning of the representations by dealing with multiple partial tasks simultaneously. Experimental results show that the proposed framework outperforms previously proposed hierarchical multi-label contrastive losses in fine-grain UI classification.
\end{abstract}

\section{Introduction}
\label{sec:intro}

User Interface (UI) understanding has been extensively explored in the past~\cite{rico,screen,webshop,webui,actionbert,uibert,vut,pix2struct}. These works deal with simplified versions of UIs coming from mobile apps, or more complex ones in the context of browsers. However, to our knowledge, there are no works focusing on representation learning in the context of complete computer UIs. In these contexts, several windows appear in the screen corresponding to several open applications. Computer UIs can be messy, cluttered, and they tend to have a high variance. Variations in the UI can happen because of operating systems. Additionally, application software can differ from one operating system to another not only due to different versions, but due to special configurations applied by an user. At the same time, websites contain a lot of distracting information, like images, publicity, text boxes, buttons and colors that might resemble selected text. Whereas obtaining supervision for the structure of web applications is fairly straightforward, obtaining the same level of supervision is not feasible for most computer software applications due to a lack of access to a low-level representation of the screen similar to the HTML code of a website. Besides lack of access to structural supervision, other important characteristics such as open context menus and text selections would not be present in such structure.

In particular, we are interested in automating workflow processes in an enterprise environment by observing humans perform tasks in a computer. For instance, suppose a user first opens an email, selects and copies some identifier in the text of the email, opens a web application in a browser, pastes the selection into a search component, selects and copies some information from the search, goes back to the email application, writes a reply email, pastes the information found, and sends the email. Now assume we would like to build a system that given the observed sequence of actions applied in given states, generates a program that executes this pipeline in a new email. The first step in building such system would require to understand, given an image, what the current state of the computation is. In this paper, we focus on this task. We first define a language for describing the state of computation. It includes the application the user is interacting with at some point in time, the view of that application, and the context of the interaction (e.g. selected text). Then, we captured videos of a user performing some actions in a sequence that correspond to realistic interactions with the computer. Third, we developed a synthetic data augmentation process to generate variations of the acquired videos. Finally, we developed a learning framework to acquire representations to create classification models of the state characteristics.

We propose UI Multi-task Contrastive Learning (\framework), a novel semi-supervised framework for learning labeled and unlabeled characteristics in computer UIs. As a result, our framework consists of two modules: a \textit{synthetic sample generator} in charge of creating new instances of unlabeled characteristics in the UI for training, and an \textit{embedding network} that produces representations from which we can extract both labeled and unlabeled characteristics. Experiments on our newly introduced  \dataset dataset show improvements across several metrics over baseline methods.

Our paper makes three main contributions: (1) it introduces a new UI understanding task focusing on looking at the computer screen as a state; (2) it presents a new framework to create synthetic samples of unlabeled characteristics and learn representations of the noisy inputs; and (3) it introduces a new dataset which allows research into the task from the point of view of unsupervised and semi-supervised learning.

\section{Computer User Interface Understanding}

The primary objective of this novel learning task is to stimulate the development of models that can understand a computer UI as a state. At the same time, these models should empower control systems that allow users to automate specific workflows. Due to the lack of images/videos of computer UIs with clean and complete labels, we focus in this paper on semi-supervised and unsupervised representation learning methods.

We represent the contents of a computer screen using a set of labels that allows for further building of workflow automation processes:

\begin{itemize}
    \item \textbf{Software Application:} Name of the software application on the active window in the screen. Examples are: Browser, Spreadsheet, or Document editor.

    \item \textbf{View:} Within the previous active application, the current operation being performed. Examples are: save file, show options, or main view.

    \item \textbf{Context:} Extra information about the  interaction with the current application. In this work we differentiate between three different context values:

    \begin{itemize}

        \item \textbf{Context menu:} Whether a context menu is open on the screen.
    
        \item \textbf{Selected text:} Whether there is a text selected on the screen.

        \item \textbf{None:} Neither a context menu nor a text selection appears on the screen.
    \end{itemize}
\end{itemize}

The end goal of this representation learning task is for the representations to be used to solve downstream tasks. Some of the downstream tasks future work could help address include:

\begin{itemize}
    \item \textbf{Context-aware Interaction Recognition:} Exploring techniques to recognize and interpret user interactions and intentions, allowing for more natural and efficient control of the computer.

    \item \textbf{User-Centric AI Assistance:} Focusing on the development of AI models that assist users by predicting their actions and providing recommendations within the UI.

    \item \textbf{Task Automation:} Development of automated control systems that can accomplish tasks moving seamlessly between different software applications while carrying out cognitive complex functions.
\end{itemize}

\section{Related Work}
\label{sec:related}

\textbf{UI understanding} is a challenging task. Some works have focused on UI retrieval~\cite{rico,design,swire,screen,vins,gnn} and generation~\cite{lgan,lvae,blt,ndn,ltvae}. Others have dealt with UI understanding in the context of digital device control~\cite{wob,datad,pix2act,language,instruction,webshop}. The closest works to ours have focused on providing HTML, view hierarchy or natural language descriptions of the screen and its controls for simplified web and mobile apps with access to heavy supervision~\cite{actionbert,uibert,vut,pix2struct,spotlight}. Our task is more challenging as we deal with non-simplified views of diverse software applications in a computer environment with little supervision.

\textbf{Synthetic Samples} have been proven not only useful when gathering data is not easy, but also when labeling data is expensive. Some tasks such as recognition~\cite{txtrecsynth}, detection~\cite{odsynthetic}, pose estimation~\cite{sim2real}, depth estimation~\cite{high}, optical flow~\cite{naturalistic,optical}, or alignment and restoration~\cite{sidar} have been shown to benefit from synthetic samples. Our synthetic sample generator shares similarities with the one presented by Gupta \etal~\cite{txtrecsynth}. However, our application does not need to project text or context menus using depth. At the same time, we create synthetic samples conditioned on characteristics present on our screen.

\textbf{Contrastive Learning} is an extensively researched area. From self-supervis-ed~\cite{contrastive,con3,con4,con6}, multi-modal~\cite{clip} and supervised frameworks~\cite{supcon}, we have seen performance increase across different tasks. Examples include image recognition~\cite{recon1, recon2}, object detection~\cite{od1}, image segmentation~\cite{seg1}, action recognition~\cite{ar2}, localization~\cite{al2} and segmentation~\cite{as1}. 

Recently, we have seen an increased interest on the multi-label setting of contrastive learning~\cite{multi1,multi2, multi3,multi4,hier,multi5,multi6}. Małkiński \etal~\cite{multi1} defined a multi-label contrastive framework by creating positive samples from images that shared at least one label. Hoffmann \etal~\cite{multi6} proposed a recursive version of supervised contrastive learning in which each class has a ranking of related classes through an assigned super class. Zhang \etal~\cite{hier} introduced the natural hierarchical structure of a dataset into the learning objective by enforcing a hierarchy-based weighting. 

Although our framework uses the labels' hierarchy, our goal differs from the ones in~\cite{hier,multi6}. The proposed framework does not try to enforce any hierarchical structure into the embedding. Our approach is a one-step framework which leverages multiple tasks and the hierarchy to prevent noise corruption, thus improving representations learned by our main projection head.
\begin{figure*}[!t]
  \centering
   \includegraphics[width=\linewidth]{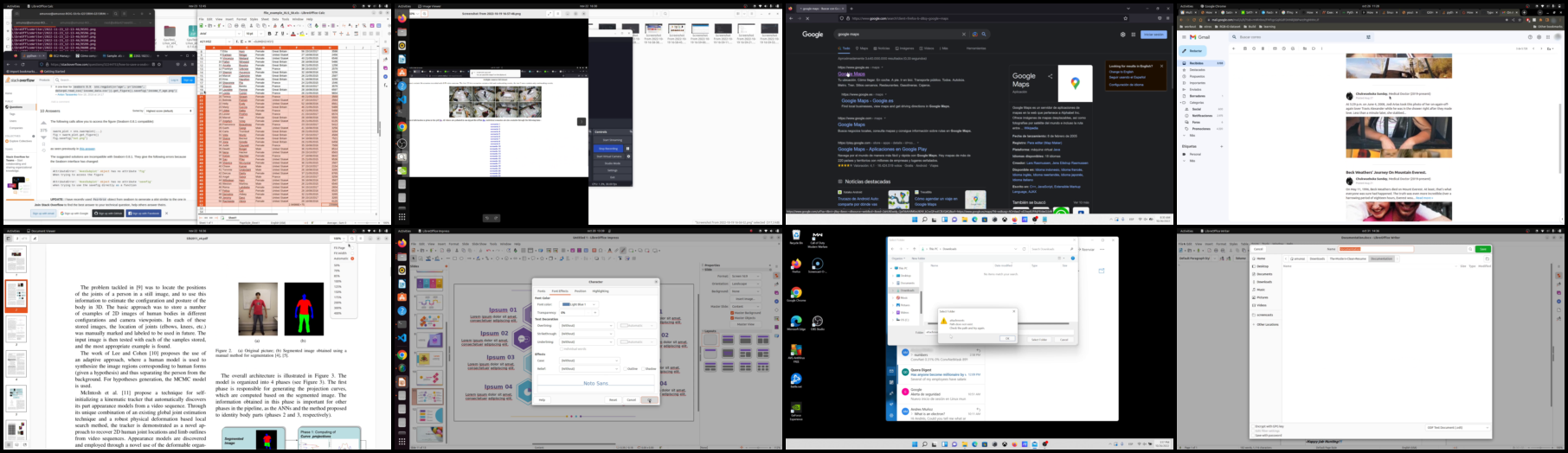}

   \caption{Sample images taken from \dataset.}
   \label{fig:datasamples_img}
\end{figure*}
\section{Dataset}
\label{sec:dataset}

As we mentioned in~\Cref{sec:related}, previous works focused on web and mobile apps, thus there are no existing datasets for UI understanding with a diverse set of computer software applications. \Cref{fig:datasamples_img} shows some samples of the computer screen in diverse settings found in the dataset.

We introduce \dataset, a novel dataset for UI understanding in the context of computer workflows. The dataset is comprised of 13894 frames were extracted from 88 videos recorded by us. \dataset includes 26 classes describing software and view and up to 78 classes describing the combination of software, view and context. It also contains labels for context menu and selected text recognition. The recorded videos are divided into train and test sets; the train set contains 9597 frames while the test set contains 4297 frames. The test set contains labels from software, view and context, whereas on the train set only software and view are labeled.


In this paper \dataset is used for semi-supervised representation learning. Nonetheless, in the future, the dataset could be useful for unsupervised representation learning, action recognition,  learning from demonstration, open-set learning and anomalous behaviour detection, etc.

In this section we describe in detail how the dataset was recorded, cleaned and labeled.

\begin{figure*}[t]
  \centering
   \includegraphics[width=\linewidth]{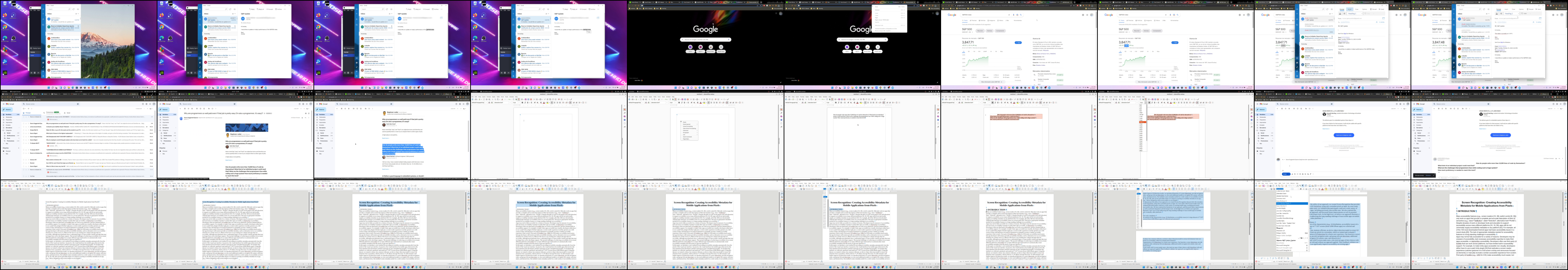}

   \caption{Samples of sequences taken from \dataset.}
   \label{fig:datasamples}
\end{figure*}

\subsection{Dataset Recording}

We acquired data through mock-up workflows. First, we recorded the mock-up workflows using OBS Studio's\footnote[1]{https://obsproject.com/} screen recorder. The workflows were recorded by a single person over the course of a month at 25 FPS with a resolution of $1920 \times 1080$. Most videos show long form coherent actions, like the ones presented on~\Cref{fig:datasamples}. Others are composed of random short-form actions performed in a specific software application.~\Cref{fig:actionlabels} shows the distribution of long-form action labels and videos with non-coherent long-form actions. As mentioned in~\Cref{sec:intro} we have a specific interest in enterprise workflow automation, thus coherent long-form actions performed follow common enterprise use cases. 



\subsection{Data Labeling}

At the time of recording, we used \textsc{xdotool}\footnote[2]{https://launchpad.net/ubuntu/trusty/+package/xdotool} to record the name of the software application on screen and the view in the software. However, tools like \textsc{xdotool} are not perfect. They tend to make mistakes because they are out of sync with the recording software, names are not standardized, and some views just return an empty string. In order to have correct and standardized labels we had to do a thorough, frame by frame, quality assurance process. At the same time, we added the labels for context only on the test set. \Cref{tab:labels} shows the distribution of the software and view labels throughout the dataset, the distribution is consistent with enterprise use of computer software. \Cref{tab:labels2} presents the distribution of the context labels in the test.\\


\begin{minipage}{\linewidth}
  \begin{minipage}[t]{0.49\linewidth}
    \centering
    \captionsetup{width=0.95\linewidth}
      \captionof{table}{\dataset software-view labels distribution.}
      \label{tab:labels}
    \resizebox{0.99\linewidth}{!}{
  \begin{tabular}{@{}cccc@{}}
    \toprule
    Software Application & View & Percentage Recorded \\
    \midrule
    File Explorer & Grid View & 3.00  \\
    File Explorer & Options  & 0.16\\
    Web Browser & Gmail  & 16.58 \\
    Web Browser & Google  & 1.78 \\
    Web Browser & Maps  & 6.58\\
    Web Browser & New Tab  & 2.27\\
    Web Browser & Save  & 0.22 \\
    Web Browser & Web Page  & 2.74\\
    Web Browser & Web PDF  & 2.50\\
    Web Browser & Options  & 1.24 \\
    Spread Sheet & Main View  & 24.85 \\
    Spread Sheet & Save  & 0.21 \\
    Spread Sheet & Options  & 0.30 \\
    Image Viewer & Main View  & 0.69\\
    Document Editor & Main View  & 19.93 \\
    Document Editor & Save  & 1.73 & \\
    Document Editor & Options & 1.12 \\
    PDFViewer & Main View  & 1.47 \\
    Terminal & Main View  & 2.52 \\
    Mail & Gmail  & 7.25 \\
    Mail & Save  & 0.11 \\
    Desktop & Main View  & 0.14 \\
    Presentation Editor & Main View  & 2.32 \\
    Presentation Editor & Save  & 0.02 \\
    Presentation Editor & Options  & 0.30 \\
    \bottomrule
  \end{tabular}}
  \end{minipage}
  \hfill
  \begin{minipage}[t]{0.49\linewidth}
   
   \captionsetup{width=0.8\linewidth}
      \captionof{table}{\dataset context labels distribution.}
      \label{tab:labels2}
      \centering
    \resizebox{0.8\linewidth}{!}{
  \begin{tabular}{@{}ccc@{}}
    \toprule
    Label & Percentage Recorded \\
    \midrule
    Context Menu & 7.26 \\
    Selected Text & 30.93 \\
    None  & 61.81 \\
    \bottomrule
  \end{tabular}}
  \begin{minipage}[b]{\linewidth}
  \centering
  \includegraphics[width=0.84\linewidth]{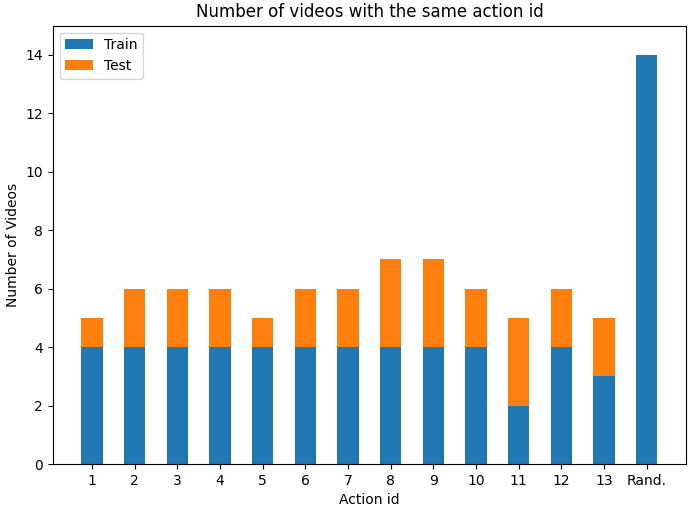}
    \captionsetup{width=0.8\linewidth}
    \captionof{figure}{Distribution of long-form actions and random actions in \dataset.}
    \label{fig:actionlabels}
  \end{minipage}
  \end{minipage}

  \end{minipage}

\section{UI Multi-task Contrastive Learning}
\label{sec:method}

\begin{figure*}[t]
  \centering
  \begin{subfigure}{0.31\linewidth}
     \includegraphics[width=0.99\linewidth]{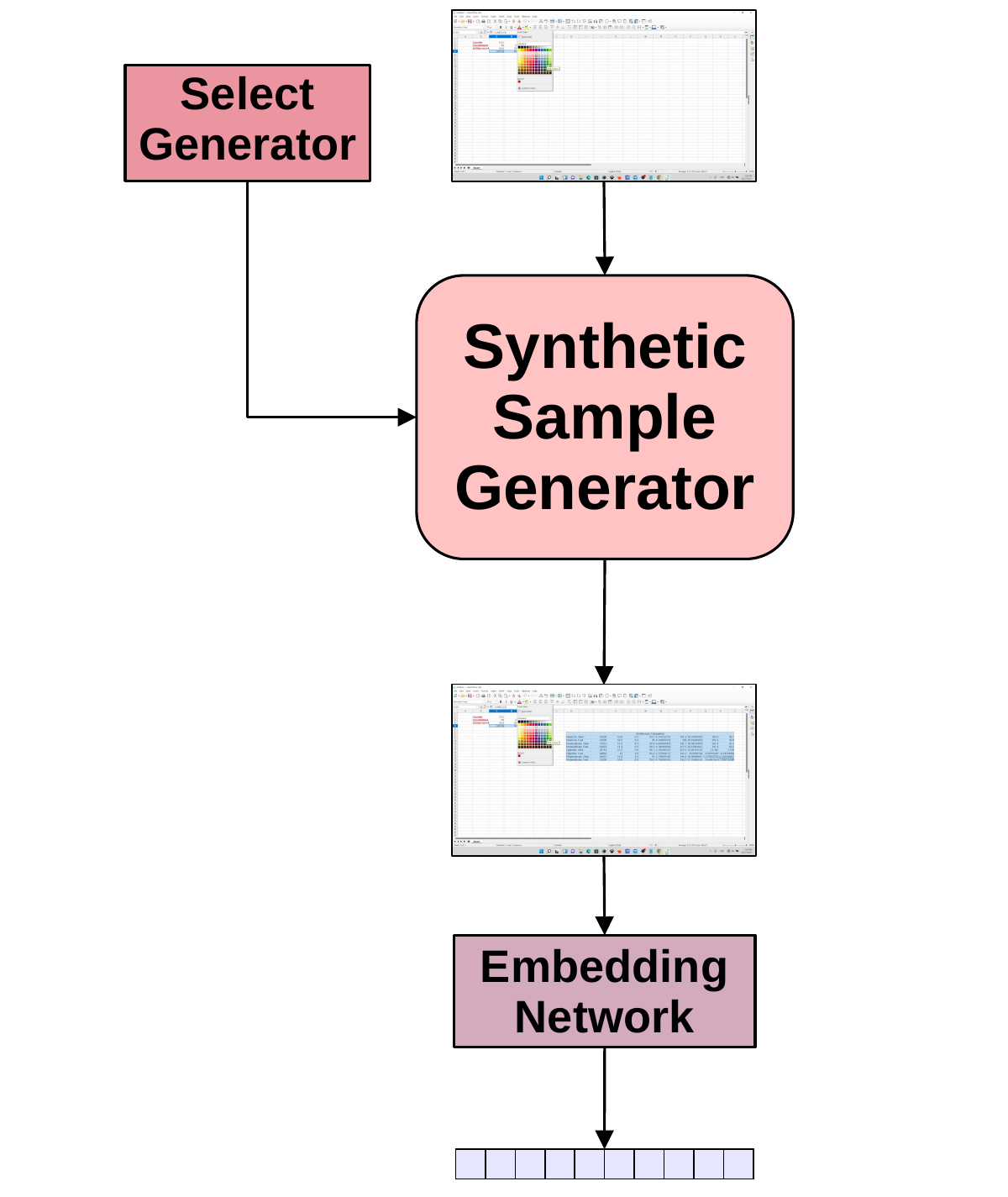}
    \caption{Full Architecture.}
    \label{fig:full}
  \end{subfigure}
  \hfill
  \begin{subfigure}{0.30\linewidth}
    \includegraphics[width=0.99\linewidth]{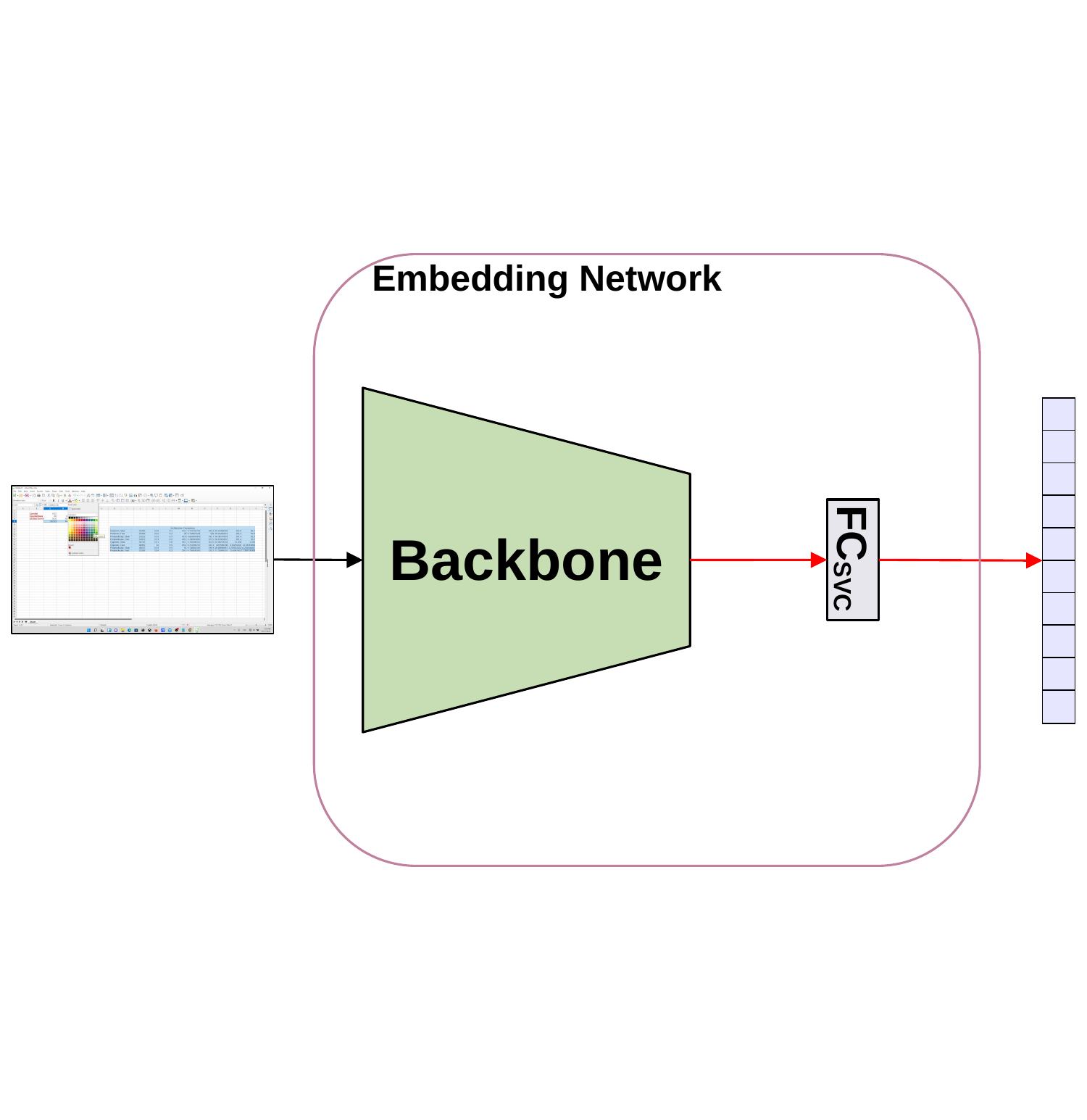}
    \caption{Single task architecture.}
    \label{fig:single}
  \end{subfigure}
  \hfill
  \begin{subfigure}{0.30\linewidth}
    \includegraphics[width=0.99\linewidth]{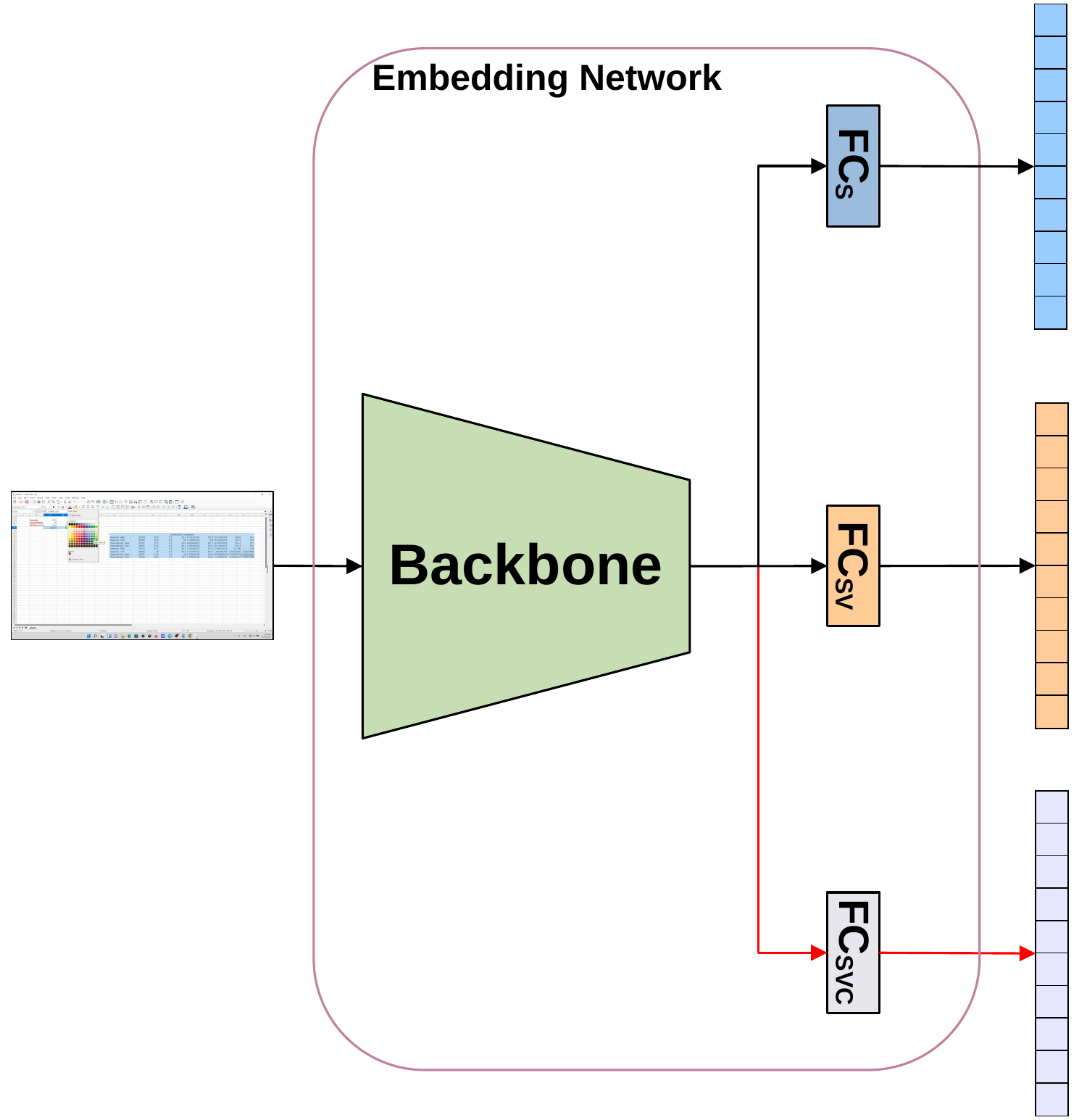}
    \caption{Multitask architecture.}
    \label{fig:multi}
  \end{subfigure}
  \caption{(a) Illustration of the full architecture of the system. (b) Shows the standard architecture of a contrastive learning task $\text{FC}_{\text{svc}}$. (c) Illustration of the proposed method, $\text{FC}_{\text{s}}$, $\text{FC}_{\text{sv}}$ and $\text{FC}_{\text{svc}}$ are the projection heads for software, software-view and software-view-context tasks, respectively. The red line symbolizes the path we are interested in for inference, while all other paths are used for training only.}
  \label{fig:arch}
\end{figure*}

Our learning approach is composed of two parts, a synthetic data generator to create the missing labels for the context menu and selected text attributes, and a supervised contrastive learning approach to learn to encode all attributes. \Cref{fig:full} illustrates the proposed architecture.

\subsection{Synthetic Data Generator}

\begin{figure*}[t]
  \centering
  \begin{subfigure}{0.45\linewidth}
     \includegraphics[width=0.98\linewidth]{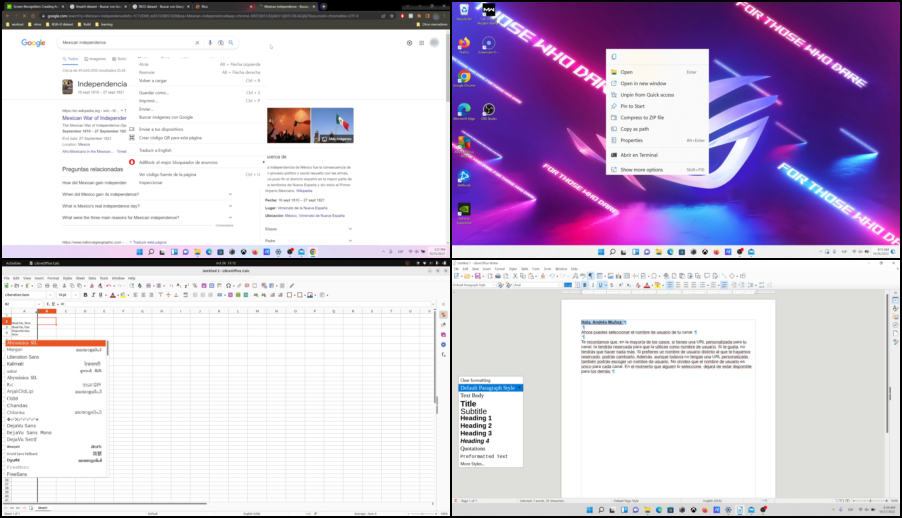}
    \caption{Synthetic context menu samples.}
    \label{fig:menus}
  \end{subfigure}
  \hfill
  \begin{subfigure}{0.45\linewidth}
    \includegraphics[width=0.98\linewidth]{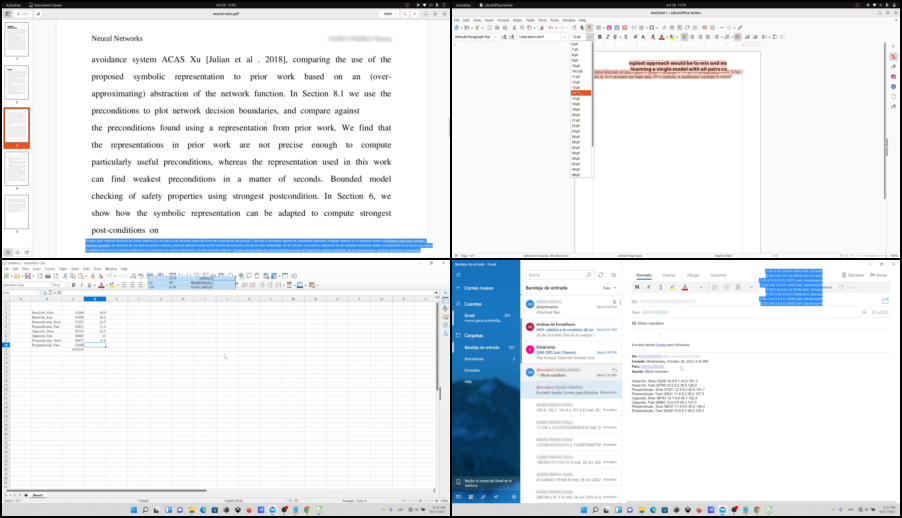}
    \caption{Synthetic selected text samples.}
    \label{fig:select}
  \end{subfigure}
  \caption{Synthetic Generator samples. (a) Shows a few samples from the synthetic menu generator. (b) Presents samples generated by the selected text generator.}
  \label{fig:synth}
\end{figure*}

In order to help the labeling process, 
we devised a synthetic sample generator which simulates the presence of either a context menu or selected text on a real image sampled from the training dataset. Since we do not have labels of those classes, we add the simulated instances irrespective of them already being naturally present on the image. \Cref{fig:synth} shows samples generated by our synthetic sample generator.

The synthetic data generator is divided into two independent generators: \textbf{Context Menu Generator} and \textbf{Selected Text Generator}. We choose between generators by sampling from a uniform distribution. Synthetic instances were added to $66.66\%$ of the dataset.


\textbf{Context Menu Generator:} We collected context menus to create a data-base by opening different drop-down menus, search bars and doing right click in various software applications on Windows 11 and Ubuntu 22.04. Then, a screenshot was taken, and we cropped to fit only the context menu.

Synthetic datapoints are generated by sampling an image from the dataset and sampling a context menu from our collected database conditioned on the software class our sampled image presents. Then, the context menu images are resized to fit within the original image's size if necessary. Finally, the context menu is placed at a random location in the image.


\textbf{Selected Text Generator:} Creating synthetic instances of selected text works in a similar fashion as generating synthetic instances of context menus. We created a database of instances of selected text, but in this case we sourced it from the training split of our dataset. As in the context menu generator, we crop the selections from their source images leaving only the selection.

The generator works by sampling an image from the dataset, and a selected text sample from the collected database. Since we have a limited set of selections, and our samples in the database are long enough to support it, we apply random cropping along the width of the selection image and random horizontal flip to create more selection images and prevent overfitting to the collected samples. Finally, the sampled selection is placed at a random location in the image.
\begin{figure}[t]
  \centering
  \includegraphics[width=0.95\linewidth]{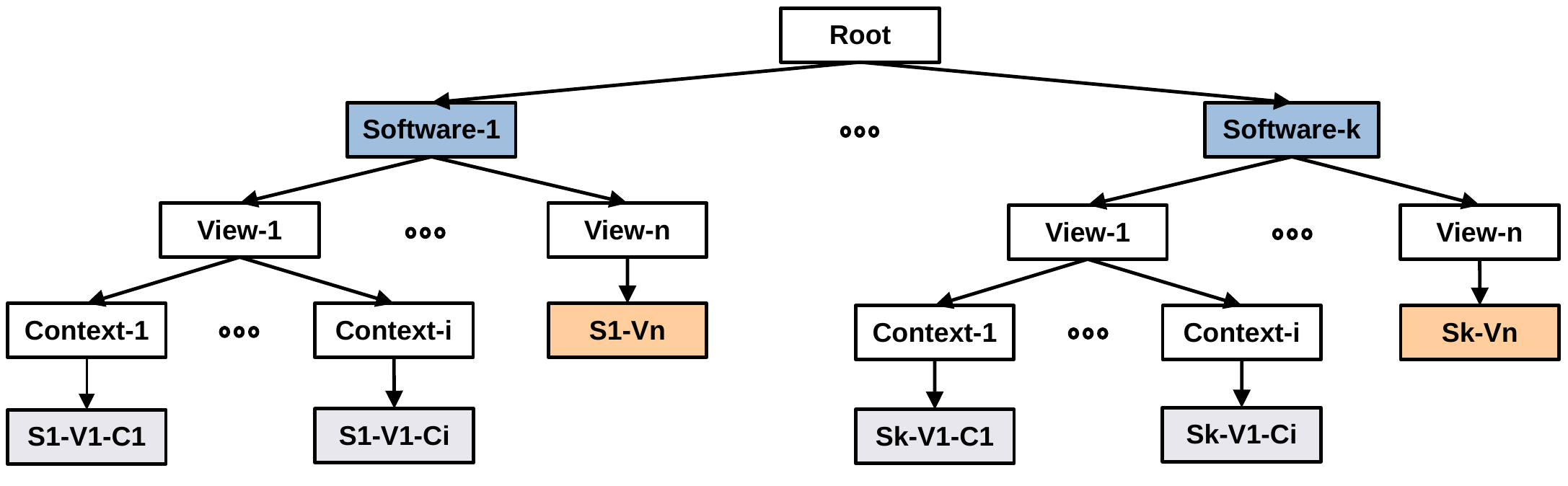}

    \caption{Label hierarchy. The leaves in purple illustrate the software-view-context hierarchical chain label used on train and test, while the orange leaves represent the software-view sub chain.}
   \label{fig:hier}
\end{figure}

\subsection{Contrastive Learning}

Supervised contrastive learning (Khosla \etal~\cite{supcon}) is an extension of the original self-supervised method introduced by Chen \etal~\cite{contrastive}. Similar to other supervised metric learning approaches~\cite{triplet,multi}, supervised contrastive learning looks to push data points of the same class together, while driving data points of different classes away. In general terms, let $i$ be the index of an image in a set of images $A$, which we call anchors, and $p$ the index of a positive image pair in a set of positive pairs $P$. Positive pairs are images which belong to the same class as our anchor $i$. Supervised contrastive learning defines the objective as:

\begin{equation}
  L^{sup} = \sum\limits_{i \in A} \frac{1}{|P_{i}|} \sum\limits_{p \in P_{i}} -\text{log} \frac{\text{exp}(f_{i} \cdot f_{p} / \tau)}{\sum\limits_{a \in A \setminus i} \text{exp}(f_{i} \cdot f_{a} / \tau)}
  \label{eq:supcon}
\end{equation}
where $f_i$, $f_a$ and $f_p$ are feature vector representations of the images and $\tau$ is a temperature parameter.

\subsection{Multi-task Contrastive Learning}
\label{sec:shl}

We adopt a similar view to labels as Zhang \etal~\cite{hier}, in which each sample has multiple labels that are dependent on one another. In some cases, such dependencies are better represented as a tree where each of its levels is a hierarchical level. Our dataset's label hierarchies are illustrated in \Cref{fig:hier}.

In contrast to Zhang \etal, we do not try to enforce the hierarchical structure into our representation, but use the shared information at different levels of the hierarchical chain to prevent the corruption of the representation due to noise introduced by our synthetic samples. In order to do this, we propose a contrastive learning framework using multi-task representation learning. Through multi-task learning, we take advantage of the software and software-view levels of the hierarchical chain to create representations which aid the learning of the full software-view-context chain. Since the \dataset dataset contains three hierarchy levels, named $s$, $sv$ and $svc$, our final architecture  (\Cref{fig:multi}) contains three independent projection heads $\text{FC}_{s}$, $\text{FC}_{sv}$ and $\text{FC}_{svc}$ tasked with learning relationships at a specific hierarchical level.  The proposed framework is optimized using a multi-task objective which we call Split Hierarchy Loss (SHL):

\begin{equation}
  L_{SHL} = \sum\limits_{l \in L} \lambda_{l}\, L^{sup}_{l}
  \label{eq:loss}
\end{equation}
where $L = \{s,sv,svc\}$, $\lambda_{l}$ are the weighting factors, and  $L^{sup}_{l}$ the contrastive loss at level $l \in L$ (\Cref{fig:hier}).

Considering the hierarchy level $svc$ contains the full information about the images, we drop $\text{FC}_{s}$ and $\text{FC}_{sv}$ during inference, using only $\text{FC}_{svc}$ for our downstream tasks.

\section{Experiments}

\subsection{Datasets and Evaluation}

We evaluate the model on our \dataset dataset. We train only on the pre-defined recorded frames train set. 



Since we are approaching the problem from a representation learning perspective, we will be evaluating the quality of the learned features using the following metrics:

\begin{itemize}

    \item \textbf{AMI}: Chance adjusted mutual information score. Given the ground truth labels and some label assignment by some clustering algorithm, AMI measures the agreement between the two assignments, ignoring permutations.
    
    \item \textbf{Precision@1}: In our retrieval experiments it is equivalent to K-NN accuracy when $k=1$. In all experiments, it is equivalent to top-1 accuracy, using standard classification with cross entropy loss.

    \item \textbf{R-Precision}: Let $R$ be the total number of references on a database to the same class as an image query $q \in Q$. We find the $R$ nearest neighbours to $q$. Out of the set $R$, $r$ is the number of those references that belong to the same class as $q$.

    \item \textbf{mAP@R}: Mean average precision at some recall level $R$.
\end{itemize}

Metrics here are calculated for the full chain $svc$. Experiments on the partial chains $s$ and $sv$ are found on the supplementary material. All metrics use only the representation learned by the $\text{FC}_{svc}$ head, except metrics for \framework-NoSynth found in \Cref{tab:ablation_synth} which use the $\text{FC}_{sv}$ head. Metrics are reported as the average metric throughout the classes in the dataset. The equations to calculate the metrics are found in the supplementary material.

When testing recorded frames $svc$ classes, the test set is divided roughly in half by video id, so there is a low likelihood of 1-to-1 matching of frames in the database set and the query set. 

\begin{table*}[t]
\caption{Clustering, retrieval and classification results. The best overall performance on each metric is highlighted in blue. The best performance per hierarchy usage level is underlined and in bold. Levels refer to the number of hierarchies used to train the model.}
\label{tab:resutls}
    \centering
    \resizebox{0.8\textwidth}{!}{
    \begin{tabular}{c|c|c|c|c|c|c|c}
        \hline
        \multirow{2}{*}{Method} & \multirow{2}{*}{Loss} & \multirow{2}{*}{Levels} & \multirow{2}{*}{Emb. dim} & \multicolumn{4}{c}{Software-View-Context}\\ 
        \cline{5-8}
        & & & & AMI  & R-Prec. & mAP@R & Prec@1\\
        \hline
        \midrule
        ST-Classification & \textbf{Cross Entropy} & 1 & -- & -- & -- & -- & 53.55\\
        \hline
        \multirow{3}{*}{\nframework} & \multirow{3}{*}{\textbf{SVC}} & \multirow{3}{*}{1} & $1\times128$ & 61.87 & 57.80 & 53.49 & 74.20\\
        
        &  & & $1\times256$ & 66.04 & \underline{\textbf{60.20}} & \underline{\textbf{54.07}} & 75.16\\
        
        &  & & $1\times384$ & \underline{\textbf{66.22}} & \underline{\textbf{60.20}} & 53.01 & \underline{\textbf{77.03}}\\
        \midrule

        MT-Classification & \textbf{Cross Entropy} & 2 & -- & -- & -- & -- & 56.06\\

        \hline
        \multirow{3}{*}{\nframework} & \multirow{3}{*}{\textbf{SHL}} & \multirow{3}{*}{2} & $1\times128$ & 67.98 & 58.61 & 52.61 & 74.20 \\
        &  & & $1\times256$ & 68.55 & 61.37 & 54.16 & 72.13 \\
        &  & & $1\times384$ & 64.46 & 60.33 & 54.58 & 71.37\\

        \hline
        \multirow{3}{*}{\nframework} & \multirow{3}{*}{\textbf{HiConMulConE}~\cite{hier}} & \multirow{3}{*}{2} & $1\times128$ & \color{blue}\underline{\textbf{70.21}} & 60.45 & 55.85 & 73.19\\
        &  & & $1\times256$ & 63.57 & 62.34 & 55.30 & 76.39\\
        &  & & $1\times384$ & 66.86 & 63.09 & 56.09 & 73.77\\

        \hline
        \multirow{3}{*}{\nframework} & \multirow{3}{*}{\textbf{RINCE}~\cite{multi6}} & \multirow{3}{*}{2} & $1\times128$ & 63.08 & 59.26 & 52.49 & 65.74\\
        &  & & $1\times256$ & 62.76 & 60.87 & 54.74 & 73.05\\
        &  & & $1\times384$ & 62.49 & 61.70 & 54.05 & 73.29\\

        \hline
        \framework & \textbf{SHL} & 2 & $2\times128$ & 69.38 & \underline{\textbf{63.65}} & \underline{\textbf{57.85}} & \color{blue}\underline{\textbf{78.16}}\\
         
        \midrule

         MT-Classification & \textbf{Cross Entropy} & 3 & -- & -- & -- & -- & 56.68\\
        
        \hline
        \multirow{3}{*}{\nframework} & \multirow{3}{*}{\textbf{SHL}} & \multirow{3}{*}{3} & $1\times128$ & 67.28 & 60.58 & 54.63 & 75.09 \\
        & & & $1\times256$ & 68.17 & 59.48 & 53.09 & 73.86\\
        &  & & $1\times384$ & 65.08 & 59.48 & 53.34 & 73.86\\

        \hline
        \multirow{3}{*}{\nframework} & \multirow{3}{*}{\textbf{HiConMulConE}~\cite{hier}} & \multirow{3}{*}{3} & $1\times128$ & 65.18 & 60.57 & 54.40 & 73.58\\
        &  & & $1\times256$ & 64.31 & 59.71 & 53.60 & 70.64\\
        &  & & $1\times384$ & 65.33 & 58.76 & 52.65 & 74.80\\

        \hline
        \multirow{3}{*}{\nframework} & \multirow{3}{*}{\textbf{RINCE}~\cite{multi6}} & \multirow{3}{*}{3} & $1\times128$ & 67.27 & 59.75 & 53.58 & 68.56\\
        &  & & $1\times256$ & 63.30 & 59.14 & 52.27 & 73.78\\
        &  & & $1\times384$ & 66.64 & 60.24 & 52.87 & 69.64 \\
        
        \hline
        \framework & \textbf{SHL} & 3 & $3\times128$ &\underline{\textbf{67.56}} & \color{blue}\underline{\textbf{65.18}} & \color{blue}\underline{\textbf{59.11}} & \underline{\textbf{75.49}} \\

    \bottomrule
    \end{tabular}}
\end{table*}

\subsection{Model Configurations}

In order to test the effectiveness of our approach we tested the following models:

\textbf{\nframework:} All models under this name follow the single task architecture seen in \Cref{fig:single}. We trained single task models using different losses. SVC refers to a single contrastive loss using the $svc$ hierarchy label. HiConMulConE is the hierarchical contrastive loss introduced in~\cite{hier}. RINCE refers to the Ranking InfoNCE loss introduced by Hoffmann \etal~\cite{multi6}. SHL is the loss introduced in \Cref{sec:shl}. The models specified have, when possible, versions using 1, 2 and 3 levels of the hierarchy. We assigned $\lambda_{sv} = 0.5$, $\lambda_{svc} = 0.5$ to SHL models trained on 2 hierarchy levels, while  $\lambda_{s} = 0.2$, $\lambda_{sv} = 0.4$ and $\lambda_{svc} = 0.4$ were chosen for models trained on 3 hierarchy levels.

\textbf{\framework:} All models tested under this name follow the multi task architecture seen in \Cref{fig:multi}. Models are trained using SHL loss, and only 2 and 3 hierarchy level models are available. The same values as above were used for $\lambda_{s}$, $\lambda_{sv}$ and $\lambda_{svc}$ on experiments with 2 and 3 hierarchical levels.
\begin{figure*}[!t]
  \centering
  \includegraphics[width=0.95\linewidth]{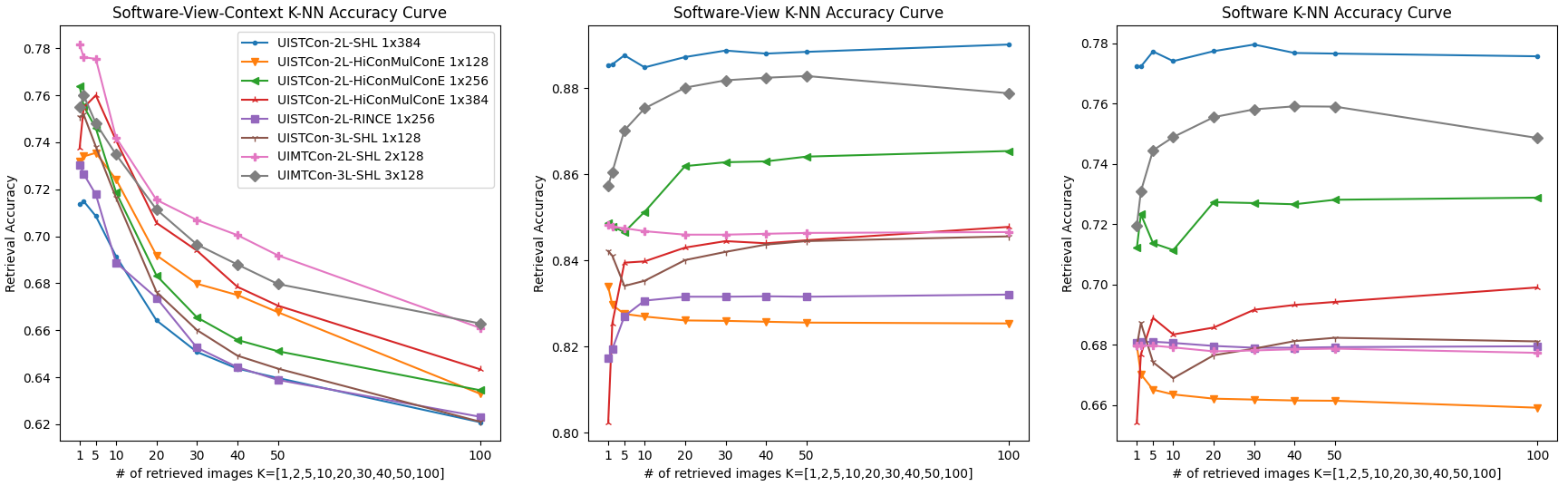}
   \caption{Retrieval results at different chain levels.}
   \label{fig:accfig}
\end{figure*}

\begin{table}[b]
    \caption{(a) Synthetic samples ablation study. (b) \framework performance across architectures.}
    \begin{subtable}{.6\linewidth}
      \centering
        \resizebox{0.95\textwidth}{!}{
  \begin{tabular}{c|c|c|c|c|c|c}
        \hline
        \multirow{2}{*}{Method} & \multirow{2}{*}{Heads} &\multirow{2}{*}{Emb. dim} & \multicolumn{4}{c}{Software-View-Context}\\
        \cline{4-7}
        & & & AMI  & R-Prec. & mAP@R & Prec@1\\
        \hline
        \midrule
         \framework-NoSynth & $\{s,sv\}$ & $2\times128$ & 62.44 & 55.02 & 49.38 & 70.54\\
         \framework & $\{s,sv,svc\}$ & $3\times128$ & \color{blue}\underline{\textbf{67.56}} & \color{blue}\underline{\textbf{65.18}} & \color{blue}\underline{\textbf{59.11}} & \color{blue}\underline{\textbf{75.49}}\\
        \bottomrule
    \end{tabular}}
        \caption{}
        \label{tab:ablation_synth}
    \end{subtable}%
    \begin{subtable}{.4\linewidth}
      \centering
        \resizebox{\textwidth}{!}{
  \begin{tabular}{c|c|c|c|c|c}
        \hline
        \multirow{2}{*}{Backbone} &\multirow{2}{*}{Emb. dim} & \multicolumn{4}{c}{Software-View-Context}\\
        \cline{3-6}
        & & AMI  & R-Prec. & mAP@R & Prec@1\\
        \hline
        \midrule
         ResNet18 & $3\times128$ & \color{blue}\underline{\textbf{70.63}} & 61.51 & 56.55 & \color{blue}\underline{\textbf{76.55}}\\
         ResNet34 & $3\times128$ & 63.48 & 63.96 & 58.14 & 73.96\\
         ResNet50 & $3\times128$ & 67.56 & \color{blue}\underline{\textbf{65.18}} & \color{blue}\underline{\textbf{59.11}} & 75.49\\
        \bottomrule
    \end{tabular}}
        \caption{}
        \label{tab:ablation_arch}
    \end{subtable} 
\end{table}

\subsection{Training Details}

Small details in the image are important for the task in hand. At the same time, essential features can be found in different parts of the image. Thus, they make pre-processing a critical part of the pipeline. In order to avoid spoiling or missing features entirely, we cannot resize to small image sizes nor apply random cropping. Hence, images are resized to 448×256×3, and then, small random brightness perturbations and horizontal flip are applied as augmentations. The backbone of the encoding network is an ImageNet~\cite{imagenet} pre-trained ResNet50~\cite{resnet}. We finetune the entire network using the Adam optimizer~\cite{adam} for 200 epochs with a choice of learning rate of $10^{-4}$, $\beta_{1}$ and $\beta_{2}$ are set to 0.9 and 0.999 respectively, the batch size to 64 and $\tau$ to 0.1. All models were trained using the same pre-processing. RINCE~\cite{multi6} and HiConMulConE~\cite{hier} are trained using the methods laid out in their original papers.

\subsection{Results}

Now, we analyze the performance of the models on the \dataset. Since there are no previous works solving this task, it is difficult to compare to previous works. Thus, our results show an ablation study and a comparison to previously proposed similar loss functions.


The retrieval experiments involve a non-parametric classification model in which we find the most relevant reference points given a query using cosine similarity as a distance metric. Meanwhile, clustering experiments used K-Means to calculate label assignments for AMI.

\textbf{Recorded Frames:} The experiments show that \framework obtains better results than the baselines in the $svc$ hierarchy level when training with 3 hierarchical labels (\framework-3L) on retrieval metrics (\Cref{tab:resutls}). \framework-3L also performs close to the top performing methods on $s$ and $sv$ levels, the results are found on the supplementary material. Furthermore, when trained on 2 hierarchical labels (\framework-2L) it produces competitive performance to \framework-3L in retrieval metrics while scoring second in clustering metrics among tested methods.

\Cref{tab:ablation_synth} shows the positive effect of the synthetic samples on all clustering, classification and, especially, on retrieval metrics where we beat \framework-NoSynth by around 10 points. Moreover, we tested \framework on different sized architectures, \Cref{tab:ablation_arch} shows that we retain most of the performance of \framework using less parameters, to the point that \framework's ResNet18 version performs better than the baseline methods using ResNet50 as a backbone.

We compare our proposed framework and baselines against supervised cross entropy on classification accuracy (precision@1). The same backbone, pre-trained weights, and training conditions were used. \framework-2L scores the highest among all methods in precision@1, as shown in \Cref{tab:resutls}, while standard classification scored the lowest by a wide margin. \Cref{fig:accfig} further supports our results by showing strong classification performance of \framework at high $k$ numbers for both \framework-3L and \framework-2L at $svc$ level, while \framework-3L is able to perform comparatively to the best performing model at $s$ and $sv$ levels without $\text{FC}_{svc}$ being directly exposed to those labels. Additionally, by observing the results presented in \Cref{tab:resutls}, we can see that classification and retrieval metrics for RINCE~\cite{multi6}, HiConMulConE~\cite{hier} and SHL tend to go down or make small gains, compared to \framework, when the software hierarchical level is added to the objective function. Experiments presented in \Cref{tab:nosynth,tab:resutls2}, in the supplementary material, show that this is likely due to noise introduced by the synthetic samples, this points to \framework being robust to noise.

\section{Conclusions}

Computer User Interface understanding is an important task for the creation of automated systems at scale in companies. We present the \framework\ framework for semi-supervised computer UI understanding. Experiments on our new dataset \dataset\ show that our framework improves the quality of the representations learned compared to evaluated baselines on in-distribution samples, while it performs comparatively on OOD samples. Our new dataset provides a platform for further research on unsupervised and semi-supervised approaches on UI representations.

\section{Acknowledgments}
This paper was prepared for informational purposes by the Artificial Intelligence Research group of JPMorgan Chase \& Co and its affiliates (“J.P. Morgan”) and is not a product of the Research Department of J.P. Morgan. J.P. Morgan makes no representation and warranty whatsoever and disclaims all liability, for the completeness, accuracy or reliability of the information contained herein. This document is not intended as investment research or investment advice, or a recommendation, offer or solicitation for the purchase or sale of any security, financial instrument, financial product or service, or to be used in any way for evaluating the merits of participating in any transaction, and shall not constitute a solicitation under any jurisdiction or to any person, if such solicitation under such jurisdiction or to such person would be unlawful.  

© 2023 JPMorgan Chase \& Co. All rights reserved


%
%
\bibliographystyle{splncs04}
\bibliography{main}
\clearpage

\appendix




\section{Rest of Hierarchy Experiments}



\Cref{tab:resutls2} and \Cref{tab:oodres2} show results at hierarchy levels $s$ and $sv$. As mentioned in the main paper, only $\text{FC}_{svc}$ is used to extract the representations used to calculate the metrics. 


Experiments in \Cref{tab:resutls2} show \nframework-3L has a competitive performance to the \nframework-SHL at $sv$ level. Both \framework-3L and \framework-2L perform in line with most of other methods at the $s$ level at their respective hierarchy level usage group. 

\begin{table*}[h]
\caption{Clustering, retrieval and classification results at the $s$ and $sv$ hierarchy levels. The best overall performance on each metric is highlighted in blue. The best performance per hierarchy usage level is underlined and in bold. Levels refer to the number of hierarchies used to train the model.}
\label{tab:resutls2}
    \centering
    \resizebox{\textwidth}{!}{
    \begin{tabular}{c|c|c|c|c|c|c|c|c|c|c|c}
        \hline
        \multirow{2}{*}{Method} & \multirow{2}{*}{Loss} & \multirow{2}{*}{Levels} & \multirow{2}{*}{Emb. dim} & \multicolumn{4}{c|}{Software} & \multicolumn{4}{c}{Software-View} \\ 
        \cline{5-12}
        & & & & AMI & R-Prec. & mAP@R & Prec@1 & AMI  & R-Prec. & mAP@R & Prec@1 \\
        \hline
        \midrule
        \multirow{3}{*}{\nframework} & \multirow{3}{*}{\textbf{SVC}} & \multirow{3}{*}{1} & $1\times128$ & 64.59 & \underline{\textbf{63.49}} & 59.44 & 67.96 & \underline{\textbf{74.66}} & \underline{\textbf{84.21}} & 82.83 & \underline{\textbf{84.91}}\\
        
        &  & & $1\times256$ & 64.45 & 59.45 & 57.42 & 64.79 & 74.65 & 81.66 & 81.00 & 79.37\\
        
        &  & & $1\times384$ & \underline{\textbf{66.03}} & 63.21 & \underline{\textbf{62.12}} & \underline{\textbf{70.32}} & 72.73 & 84.02 & \underline{\textbf{83.79}} & 84.12\\
        \midrule

        \multirow{3}{*}{\nframework} & \multirow{3}{*}{\textbf{SHL}} & \multirow{3}{*}{2} & $1\times128$ & 58.07 & 61.61 & 58.96 & 67.96 & 75.22 & 82.98 & 81.87 & 83.35 \\
        &  & & $1\times256$ & 60.74 & 59.97 & 58.87 & 66.59 & 75.59 & 83.08 & 83.01 & 83.19 \\
        &  & & $1\times384$ & 62.34 & \underline{\textbf{72.70}} & \underline{\textbf{70.96}} & \color{blue}\underline{\textbf{77.23}} & \color{blue}\underline{\textbf{80.51}} & \color{blue}\underline{\textbf{88.65}} & \color{blue}\underline{\textbf{88.10}} & \color{blue}\underline{\textbf{88.53}}\\

        \hline
        \multirow{3}{*}{\nframework} & \multirow{3}{*}{\textbf{HiConMulConE}~\cite{hier}} & \multirow{3}{*}{2} & $1\times128$ & 61.49 & 57.66 & 56.60 & 67.97 & 73.89 & 82.44 & 82.33 & 83.40\\
        &  & & $1\times256$ & \underline{\textbf{63.76}} & 68.25 & 64.99 & 71.21 & 74.88 & 86.18 & 84.86 & 84.87\\
        &  & & $1\times384$ & 63.18 & 63.75 & 60.59 & 65.43 & 77.32 & 84.51 & 83.20 & 80.25\\

        \hline
        \multirow{3}{*}{\nframework} & \multirow{3}{*}{\textbf{RINCE}~\cite{multi6}} & \multirow{3}{*}{2} & $1\times128$ & 64.11 & 54.49 & 53.00 & 63.20 & 72.08 & 81.03 & 80.83 & 81.70\\
        &  & & $1\times256$ & 60.56 & 60.88 & 59.29 & 68.06 & 74.90 & 82.82 & 82.17 & 81.73\\
        &  & & $1\times384$ & 60.30 & 59.01 & 57.69 & 64.79 & 76.44 & 82.21 & 81.92 & 78.70\\

        \hline
        \framework & \textbf{SHL} & 2 & $2\times128$ & 63.46 & 61.31 & 59.80 & 67.98 & 74.51 & 84.34 & 84.15 & 84.82\\
         
        \midrule

        \multirow{3}{*}{\nframework} & \multirow{3}{*}{\textbf{SHL}} & \multirow{3}{*}{3} & $1\times128$ & 63.94 & 64.06 & 62.16 & 67.96 & 76.45 & 84.29 & 83.49 & 84.22 \\
        & & & $1\times256$ & 60.58 & 58.73 & 57.82 & 66.20 & 74.79 & 85.57 & 84.27 & 84.14\\
        &  & & $1\times384$ & 62.31 & 62.00 & 59.93 & 69.07 & 74.77 & 83.31 & 82.65 & 85.53\\

        \hline
        \multirow{3}{*}{\nframework} & \multirow{3}{*}{\textbf{HiConMulConE}~\cite{hier}} & \multirow{3}{*}{3} & $1\times128$ & 65.48 & 67.85 & 67.06 & 67.96 & 74.55 & 82.71 & 81.97 & 82.70\\
        &  & & $1\times256$ & 64.01 & 63.58 & 62.92 & 63.20 & \underline{\textbf{80.90}} & 81.54 & 80.95 & 81.08\\
        &  & & $1\times384$ & 64.61 & 64.98 & 64.10 & 66.06 & 79.30 & 81.68 & 81.36 & 81.68\\

        \hline
        \multirow{3}{*}{\nframework} & \multirow{3}{*}{\textbf{RINCE}~\cite{multi6}} & \multirow{3}{*}{3} & $1\times128$ & 64.44 & \color{blue}\underline{\textbf{75.46}} & \color{blue}\underline{\textbf{74.75}} & \underline{\textbf{75.23}} & 74.32 & 86.30 & 86.19 & \underline{\textbf{86.43}}\\
        &  & & $1\times256$ & 65.45 & 63.85 & 63.26 & 63.20 & 76.07 & 81.43 & 81.37 & 81.57\\
        &  & & $1\times384$ & \color{blue}\underline{\textbf{77.64}} & 66.71 & 66.13 & 66.06 & 77.65 & 83.08 & 83.04 & 83.05\\
        
        \hline
        \framework & \textbf{SHL} & 3 & $3\times128$ & 59.81 & 69.31 & 66.86 & 71.96 & 77.53 & \underline{\textbf{87.22}} & \underline{\textbf{86.42}} & 85.74\\

    \bottomrule
    \end{tabular}}
\end{table*}

\section{Noise Influence Experiments}

Experiments in \Cref{tab:nosynth} seek to show the negative effect that noisy synthetic samples have on learning multi-label representations in a single embedding space. Since we are removing the synthetic samples, we lose context labels at training, thus we train only with the $s$ and $sv$ hierarchical levels.

Results show that once we remove the noise from the labels and images caused by the synthetic sample generator, all \nframework methods see a significant performance increase, when compared to results in \Cref{tab:resutls2}, to the point that most of them beat \framework. This shows that although some baseline methods, like RINCE~\cite{multi6}, were developed to deal with noisy labels, it appears noisy images still present a challenge for them. Moreover, it is visible there's a need for further research on noise resistant multi-label contrastive learning approaches.

\begin{table*}[!h]
\caption{Results on the $sv$ and $s$ hierarchical levels when training without synthetic samples. Metrics highlighted in blue have a significant performance increase w.r.t the ones trained with synthetic samples, while red signifies a significant drop. Metrics with similar performance to the ones trained with synthetic samples are not highlighted. For a fair comparison, we compare them to the experiments with the three levels found in \Cref{tab:resutls2}.}
\label{tab:nosynth}
    \centering
    \resizebox{\textwidth}{!}{
    \begin{tabular}{c|c|c|c|c|c|c|c|c|c|c|c}
        \hline
        \multirow{2}{*}{Method} & \multirow{2}{*}{Loss} & \multirow{2}{*}{Levels} & \multirow{2}{*}{Emb. dim} & \multicolumn{4}{c|}{Software} & \multicolumn{4}{c}{Software-View}\\ 
        \cline{5-12}
        & & & & AMI & R-Prec. & mAP@R & Prec@1 & AMI  & R-Prec. & mAP@R & Prec@1\\
        \hline
        \midrule
        

        \multirow{3}{*}{\nframework} & \multirow{3}{*}{\textbf{SHL}} & \multirow{3}{*}{2} & $1\times128$ & \color{blue}65.60 & \color{blue}82.57 & \color{blue}81.11 & \color{blue}77.76 & \color{blue}74.90 & \color{blue}89.36 & \color{blue}88.82 & \color{blue}87.25\\
        &  & & $1\times256$ & \color{blue}64.93 & \color{blue}74.26 & \color{blue}73.68 & \color{blue}74.34 & \color{blue}79.92 & \color{blue}86.73 & \color{blue}86.73 & \color{blue}87.10 \\
        &  & & $1\times384$ & \color{blue}65.29 & \color{blue}78.48 & \color{blue}78.48 & \color{blue}78.48 & \color{blue}79.87 & \color{blue}87.98 & \color{blue}87.97 & \color{blue}87.98\\

        \hline
        \multirow{3}{*}{\nframework} & \multirow{3}{*}{\textbf{HiConMulConE}~\cite{hier}} & \multirow{3}{*}{2} & $1\times128$ & 65.19 & \color{blue}77.09 & \color{blue}76.64 & \color{blue}76.59 & \color{blue}76.27 & \color{blue}87.49 & \color{blue}87.40 & \color{blue}87.28\\
        &  & & $1\times256$ & 64.89 & \color{blue}80.44 & \color{blue}80.01 & \color{blue}80.00 & \color{red}75.45 & \color{blue}88.83 & \color{blue}88.80 & \color{blue}88.90\\
        &  & & $1\times384$ & \color{blue}66.72 & \color{blue}78.41 & \color{blue}78.22 &\color{blue}76.24 & \color{red}76.96 & \color{blue}88.78 & \color{blue}88.72 & \color{blue}87.85\\

        \hline
        \multirow{3}{*}{\nframework} & \multirow{3}{*}{\textbf{RINCE}~\cite{multi6}} & \multirow{3}{*}{2} & $1\times128$ & 64.88 & \color{blue}77.66 & \color{blue}77.52 & \color{blue}77.34 & \color{blue}77.67 & \color{blue}88.01 & \color{blue}87.92 & \color{blue}87.77\\
        &  & & $1\times256$ & \color{blue}67.15 & \color{blue}78.14 & \color{blue}78.14 & \color{blue}78.13 & \color{blue}77.25 & \color{blue}88.39 & \color{blue}88.38 & \color{blue}88.38\\
        &  & & $1\times384$ & \color{red}66.03 & \color{blue}76.86 & \color{blue}76.63 & \color{blue}79.60 & 76.60 & \color{blue}87.14 & \color{blue}87.14 & \color{blue}87.21\\

        \hline
        \framework & \textbf{SHL} & 2 & $2\times128$ & \color{blue}62.84 & \color{blue}72.68 & \color{blue}71.73 & \color{blue}76.59 & \color{red}76.53 & 87.11 & 87.08 & \color{blue}87.26\\
        
    \bottomrule
    \end{tabular}}
\end{table*}

\begin{table*}[t]
\caption{Generalization to OOD data results.}
\label{tab:oodres2}
    \centering
    \resizebox{\textwidth}{!}{
    \begin{tabular}{c|c|c|c|c|c|c|c|c|c|c|c}
        \hline
        \multirow{2}{*}{Method} & \multirow{2}{*}{Loss} & \multirow{2}{*}{Levels} & \multirow{2}{*}{Emb. dim} & \multicolumn{4}{c|}{Software} & \multicolumn{4}{c}{Software-View}\\ 
        \cline{5-12}
        & & & & AMI & R-Prec. & mAP@R & Prec@1 & AMI  & R-Prec. & mAP@R & Prec@1\\
        \hline
        \midrule
        
        \multirow{3}{*}{\nframework} & \multirow{3}{*}{\textbf{SVC}} & \multirow{3}{*}{1} & $1\times128$ & \underline{\textbf{56.06}} & \underline{\textbf{70.14}} & \underline{\textbf{62.26}} & \underline{\textbf{59.74}} & \underline{\textbf{59.75}} & \underline{\textbf{44.34}} & \underline{\textbf{35.61}} & 23.10\\
        
        &  & & $1\times256$ & 54.78 & 59.00 & 52.87 & 55.81 & 59.16 & 35.82 & 28.87 & 22.46\\
        
        &  & & $1\times384$ & 50.43 & 57.12 & 49.03 & 57.83 & 47.20 & 39.13 & 31.79 & \underline{\textbf{28.61}}\\
        \midrule

        \multirow{3}{*}{\nframework} & \multirow{3}{*}{\textbf{SHL}} & \multirow{3}{*}{2} & $1\times128$ & 57.67 & \underline{\textbf{68.70}} & \underline{\textbf{63.68}} & 60.87 & 56.30 & 46.66 & 40.95 & 26.68\\
        &  & & $1\times256$ & 52.39 & 61.06 & 55.79 & 42.44 & 57.65 & 42.37 & 39.25 & 30.86\\
        &  & & $1\times384$ & 55.43 & 68.61 & 62.19 & 59.17 & 57.73 & 47.83 & 41.15 & 28.62\\

        \hline
        \multirow{3}{*}{\nframework} & \multirow{3}{*}{\textbf{HiConMulConE}~\cite{hier}} & \multirow{3}{*}{2} & $1\times128$ & 51.89 & 60.76 & 54.09 & 55.82 & 48.85 & 39.05 & 31.92 & 19.79\\
        &  & & $1\times256$ & 56.57 & 63.53 & 56.30 & 55.28 & 55.91 & 42.27 & 35.67 & 32.02\\
        &  & & $1\times384$ & 61.69 & 64.53 & 57.83 & 60.28 & 52.79 & 45.85 & 39.91 & 35.61\\

        \hline
        \multirow{3}{*}{\nframework} & \multirow{3}{*}{\textbf{RINCE}~\cite{multi6}} & \multirow{3}{*}{2} & $1\times128$ & 47.53 & 59.94 & 52.81 & 61.57 & 56.55 & 45.26 & 36.68 & 29.48\\
        &  & & $1\times256$ & 66.89 & 63.93 & 58.91 & \color{blue}\underline{\textbf{74.24}} & 60.69 & \underline{\textbf{49.98}} & \color{blue}\underline{\textbf{46.28}} & \color{blue}\underline{\textbf{51.42}}\\
        &  & & $1\times384$ & \underline{\textbf{67.11}} & 62.45 & 55.41 & 60.28 & \underline{\textbf{60.89}} & 44.73 & 36.84 & 30.23\\

        \hline
        \framework & \textbf{SHL} & 2 & $2\times128$ & 53.28 & 55.42 & 47.27 & 60.52 & 55.12 & 38.07 & 31.25 & 25.76\\
         
        \midrule

        \multirow{3}{*}{\nframework} & \multirow{3}{*}{\textbf{SHL}} & \multirow{3}{*}{3} & $1\times128$ & 70.06 & 68.55 & 64.73 & 66.35 & 63.73 & 48.18 & \underline{\textbf{44.71}} & 30.19\\
        & & & $1\times256$ & 52.83 & 61.03 & 49.68 & 58.76 & 57.08 & 40.99 & 33.64 & 27.78\\
        &  & & $1\times384$ & 65.56 & 69.38 & 64.57 & 58.51 & \color{blue}\underline{\textbf{68.26}} & \color{blue}\underline{\textbf{50.39}} & 43.40 & 32.08\\

        \hline
        \multirow{3}{*}{\nframework} & \multirow{3}{*}{\textbf{HiConMulConE}~\cite{hier}} & \multirow{3}{*}{3} & $1\times128$ & 69.63 & 52.94 & 59.60 & 54.19 & 57.51 & 28.76 & 24.69 & 21.05\\
        &  & & $1\times256$ & \color{blue}\underline{\textbf{73.02}} & 64.14 & 59.84 & 54.81 & 62.60 & 42.79 & 37.01 & 30.21\\
        &  & & $1\times384$ & 71.80 & \color{blue}\underline{\textbf{71.12}} & \color{blue}\underline{\textbf{66.60}} & 59.01 & 67.02 & 46.85 & 41.70 & 28.88\\

        \hline
        \multirow{3}{*}{\nframework} & \multirow{3}{*}{\textbf{RINCE}~\cite{multi6}} & \multirow{3}{*}{3} & $1\times128$ & 50.54 & 65.44 & 60.43 & \underline{\textbf{66.37}} & 43.08 & 39.84 & 31.81 & 28.32\\
        &  & & $1\times256$ & 52.48 & 59.03 & 55.05 & 59.93 & 53.42 & 28.04 & 35.14 & 27.12\\
        &  & & $1\times384$ & 47.88 & 63.10 & 59.30 & 60.81 & 54.27 & 34.38 & 40.73 & 29.29\\
        
        \hline
        \framework & \textbf{SHL} & 3 & $3\times128$ & 60.87 & 63.89 & 59.42 & 66.07 & 60.46 & 46.46 & 41.40 & \underline{\textbf{34.91}}\\

    \bottomrule
    \end{tabular}}
\end{table*}

\section{T-SNE Visualization}

Visualizing the data is helpful to further understand the quantitative results. In order to visualize the results, we project the data into two dimensions using t-sne~\cite{tsne}. On the one hand, \Cref{fig:tsne_test} shows a more clear separation of classes in \Cref{fig:tsne_mt_3l,fig:tsne_mt_2l}, which coincides with the quantitative results shown in the main paper. On the other hand, \Cref{fig:tsne_train} illustrates that models trained without synthetic samples tend to generate clusters of a higher quality (\Cref{fig:tsne_st_sv,fig:tsne_hicon_sv,fig:tsne_rince_sv}), when compared to the ones trained with synthetic samples (\Cref{fig:tsne_st_synth,fig:tsne_hicon_synth,fig:tsne_rince_synth}).

\section{Action Labels}

The long term actions present in our dataset as seen in Figure 3 of the main paper are the following:

\begin{itemize}
    \item \textbf{Action ID 1:} Select, copy and paste information from an email to search. Select, copy and paste results back into an email and send.
    \item \textbf{Action ID 2:} Open web browser, go to maps and search address.
    \item \textbf{Action ID 3:} Select, copy and paste information from an email to text document. Edit text, select, copy and paste results back into an email and send.
    \item \textbf{Action ID 4:} Save a text document.
    \item \textbf{Action ID 5:} Format a text document.
    \item \textbf{Action ID 6:} Add page count/header to a text document.
    \item \textbf{Action ID 7:} Export text document to PDF.
    \item \textbf{Action ID 8:} Select, copy and paste information from an email to spread sheet. Apply summation, select, copy and paste results back into an email and send.
    \item \textbf{Action ID 9:} Select, copy and paste information from an email to spread sheet. Apply multiplication, select, copy and paste results back into an email and send.
    \item \textbf{Action ID 10:} Select, copy and paste information from an email to spread sheet. Apply division, select, copy and paste results back into an email and send.
    \item \textbf{Action ID 11:} Select, copy and paste information from an email to spread sheet. Apply subtraction, select, copy and paste results back into an email and send.
    \item \textbf{Action ID 12:} Format a spreadsheet.
    \item \textbf{Action ID 13:} Download attachment from email and visualize it.
    
\end{itemize}

\section{Metrics}
As mentioned in the main paper, we define here formally the metrics used to score and compare quantitatively the trained methods as:

\begin{itemize}

    \item \textbf{AMI}: The score is calculated as:
    
    \begin{equation}
        \text{AMI} = \frac{\text{MI}\; E\left[\text{MI} \right]}{\text{mean}(H(U), H(V)) - \left[ \text{MI} \right]}
    \end{equation}
    where $H(U)$ and $H(V)$ are the entropy of ground truth labels $U$ and the predicted label assignment $V$ respectively and MI is the mutual information.

    \item \textbf{R-Precision}: The score is:
        \begin{equation}
            \text{R-Precision} =  \frac{1}{|Q|} \sum\limits_{q \in Q} \frac{r_q}{R_q}
            \label{eq:rprec}
        \end{equation}

    \item \textbf{mAP@R}: The score is defined as:
        \begin{equation}
            \text{mAP@R} =  \frac{1}{R} \sum\limits_{i \in R} P(i)
            \label{eq:mapr}
        \end{equation}

        \begin{equation}
            P(i) = 
            \begin{cases}
            \text{precision at } i,& \text{if the ith retrieval is correct}\\
            0,                      & \text{otherwise}
            \end{cases}
            \label{eq:preci}
    \end{equation}
\end{itemize}

\section{More Samples}

\Cref{fig:datasamples_img} shows samples taken at random from different videos of the recorded frames dataset. Sample sequences are illustrated in \Cref{fig:datasamples_seq}.




\begin{figure*}[!t]
  \centering
   \includegraphics[width=\linewidth]{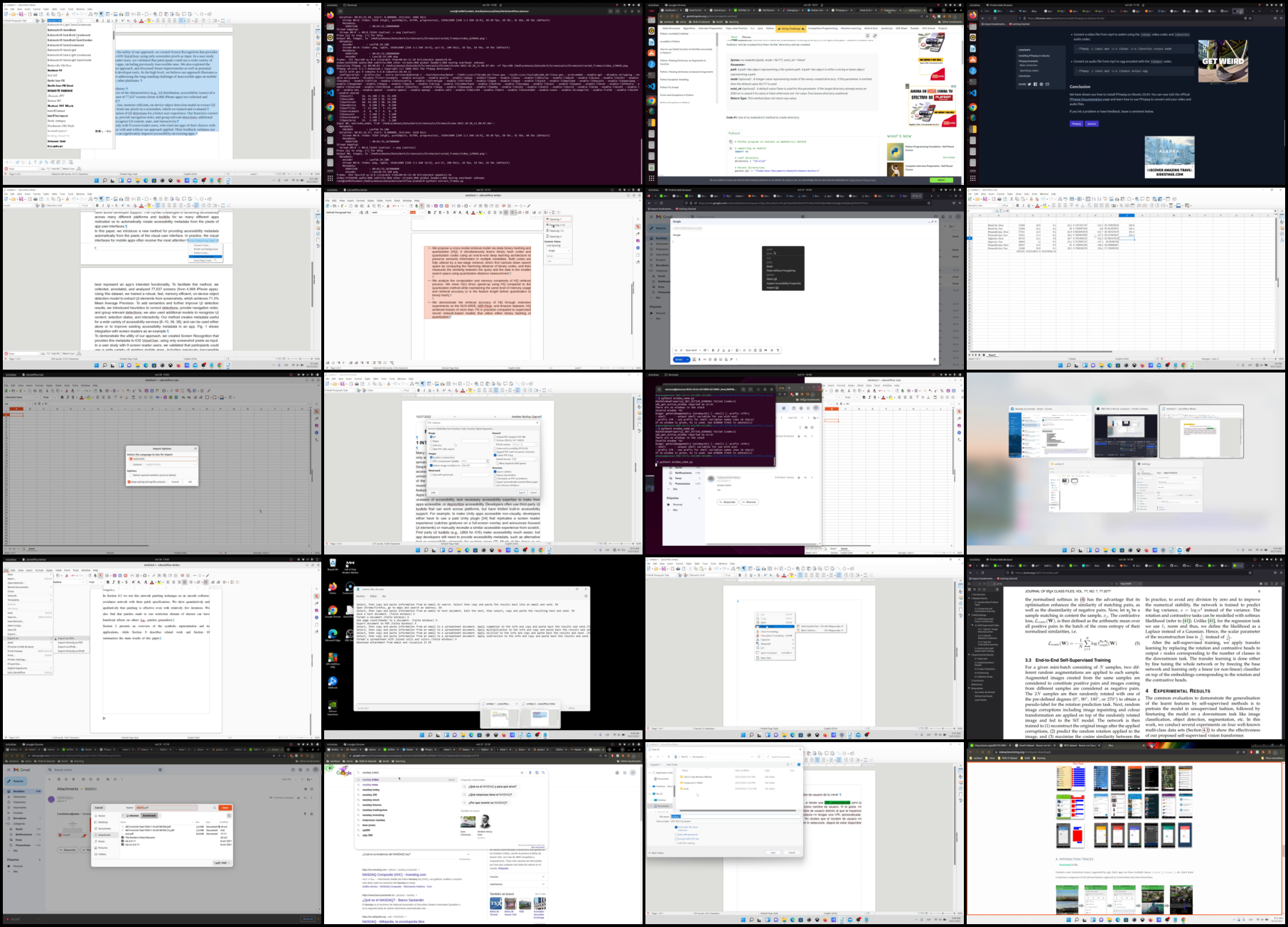}

   \caption{Sample images taken from \dataset's recorded frames set.}
   \label{fig:datasamples_img}
\end{figure*}

\begin{figure*}[!t]
  \centering
   \includegraphics[width=\linewidth]{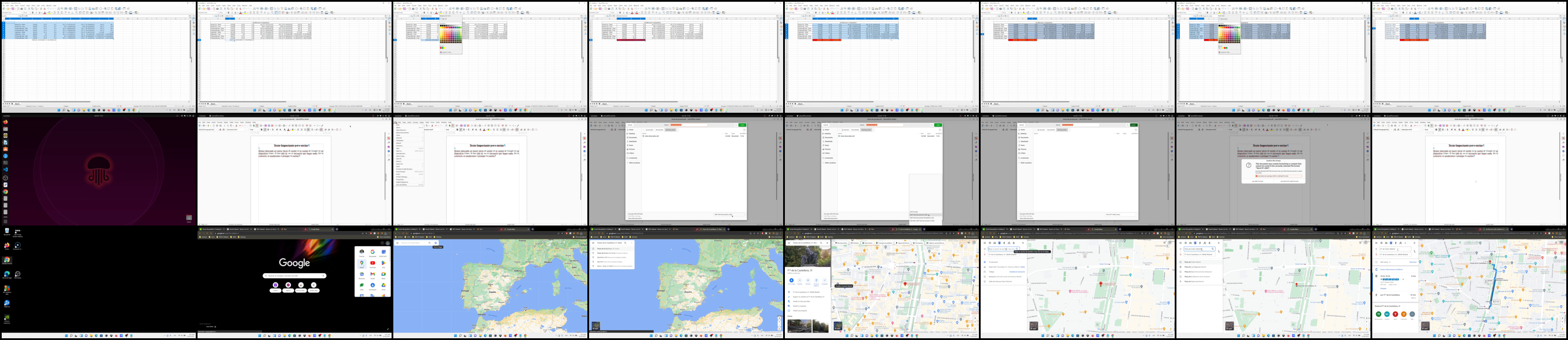}

   \caption{Sample sequences taken from \dataset's recorded frames set.}
   \label{fig:datasamples_seq}
\end{figure*}



\begin{figure*}[t]
  \centering
  \begin{subfigure}{0.32\linewidth}
     \includegraphics[width=0.99\linewidth]{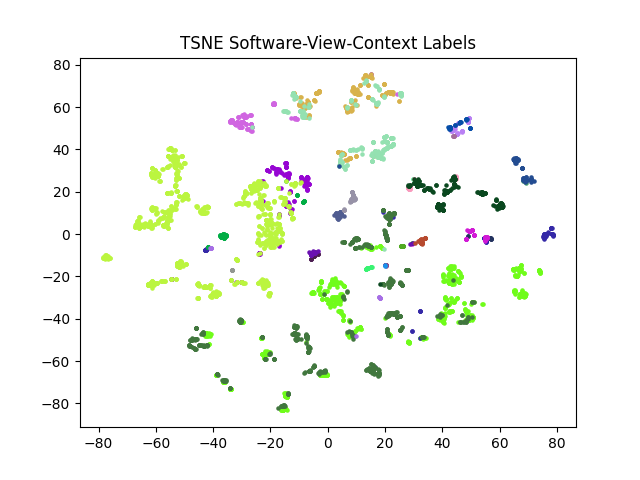}
    \caption{\framework-3L}
    \label{fig:tsne_mt_3l}
  \end{subfigure}
  \hfill
  \begin{subfigure}{0.32\linewidth}
    \includegraphics[width=0.99\linewidth]{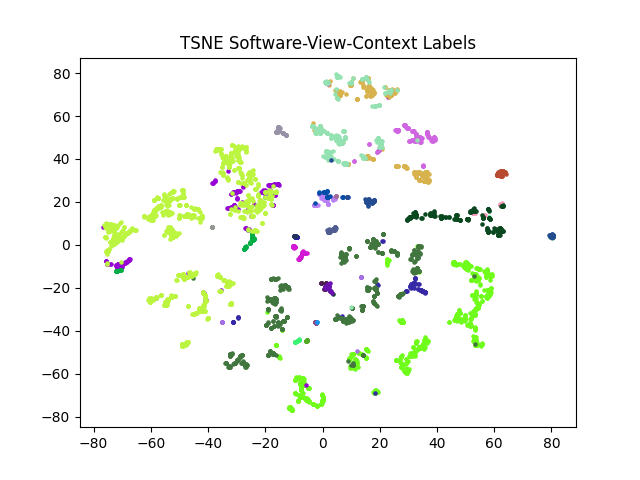}
    \caption{\framework-2L}
    \label{fig:tsne_mt_2l}
  \end{subfigure}
  \hfill
  \begin{subfigure}{0.32\linewidth}
    \includegraphics[width=0.99\linewidth]{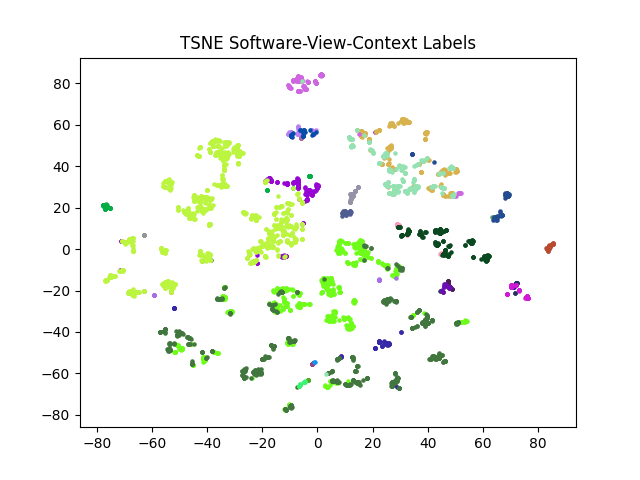}
    \caption{HiConMulConE-2L}
    \label{fig:tsne_hier_2l}
  \end{subfigure}
  \begin{subfigure}{0.32\linewidth}
     \includegraphics[width=0.99\linewidth]{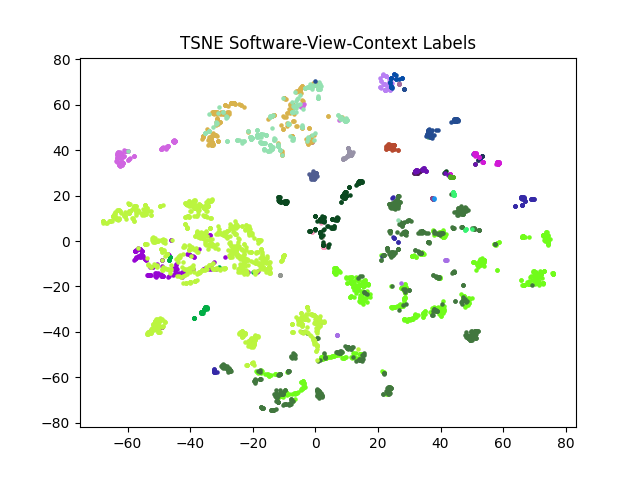}
    \caption{RINCE-2L}
    \label{fig:tsne_rince_2l}
  \end{subfigure}
  \hfill
  \begin{subfigure}{0.3\linewidth}
    \includegraphics[width=0.99\linewidth]{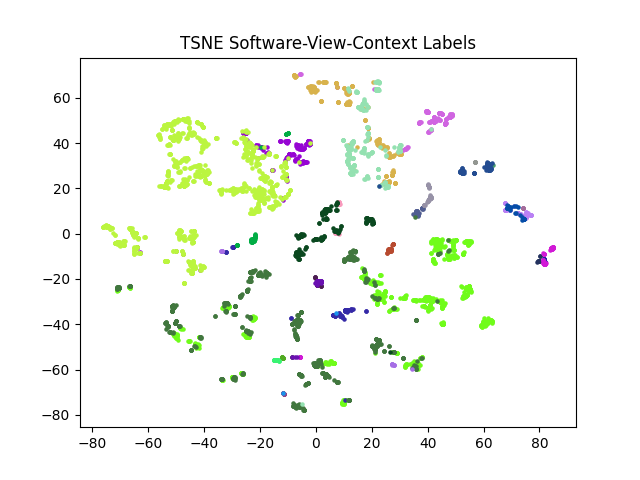}
    \caption{\nframework-SHL-1L}
    \label{fig:tsne_st_1l}
  \end{subfigure}
  \hfill
  \begin{subfigure}{0.32\linewidth}
    \includegraphics[width=0.99\linewidth]{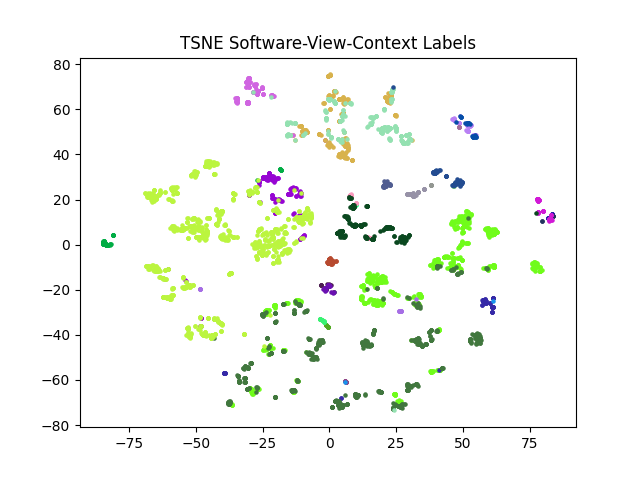}
    \caption{\nframework-SHL-3L}
    \label{fig:tsne_st_3l}
  \end{subfigure}
  \caption{TSNE projection of the recorded frames test set produced by the best mAP@R performing architectures. Colors represent the $svc$ class a datapoint belongs to.}
  \label{fig:tsne_test}
\end{figure*}

\begin{figure*}[t]
  \centering
  \begin{subfigure}{0.32\linewidth}
     \includegraphics[width=0.99\linewidth]{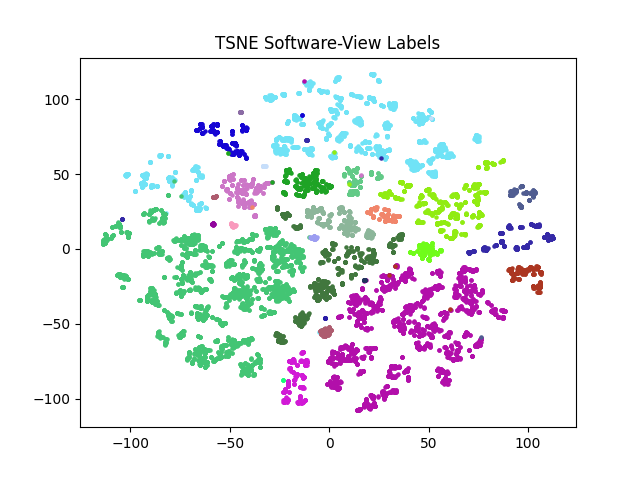}
    \caption{\nframework-SHL}
    \label{fig:tsne_st_synth}
  \end{subfigure}
  \hfill
  \begin{subfigure}{0.32\linewidth}
    \includegraphics[width=0.99\linewidth]{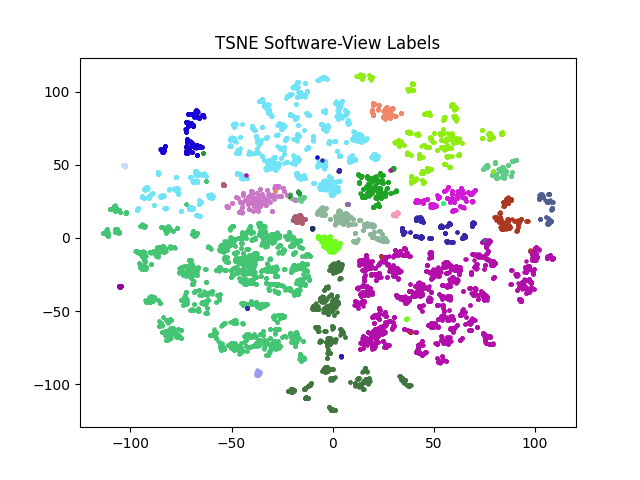}
    \caption{HiConMulConE}
    \label{fig:tsne_hicon_synth}
  \end{subfigure}
  \hfill
  \begin{subfigure}{0.32\linewidth}
    \includegraphics[width=0.99\linewidth]{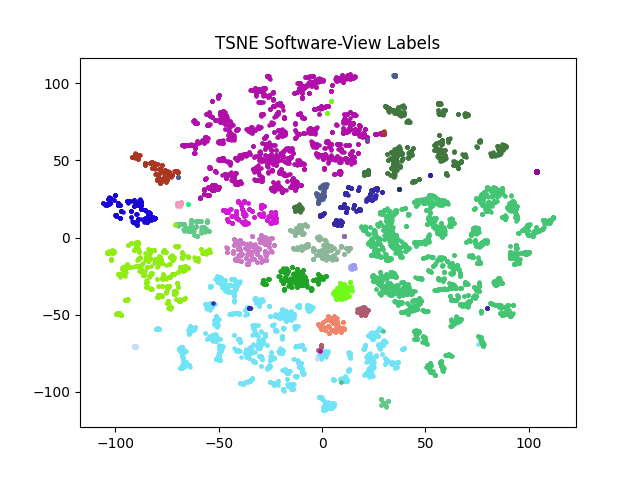}
    \caption{RINCE}
    \label{fig:tsne_rince_synth}
  \end{subfigure}
  \begin{subfigure}{0.32\linewidth}
     \includegraphics[width=0.99\linewidth]{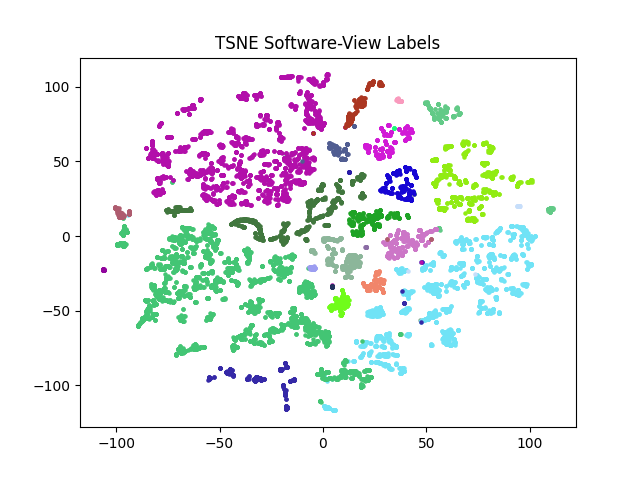}
    \caption{\nframework-SHL}
    \label{fig:tsne_st_sv}
  \end{subfigure}
  \hfill
  \begin{subfigure}{0.32\linewidth}
    \includegraphics[width=0.99\linewidth]{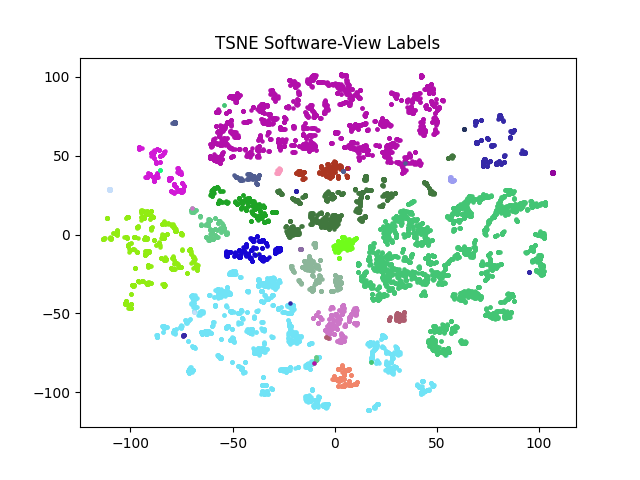}
    \caption{HiConMulConE}
    \label{fig:tsne_hicon_sv}
  \end{subfigure}
  \hfill
  \begin{subfigure}{0.32\linewidth}
    \includegraphics[width=0.99\linewidth]{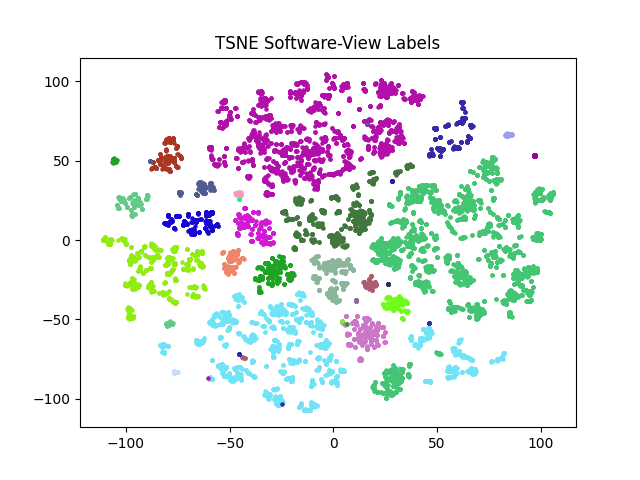}
    \caption{RINCE}
    \label{fig:tsne_rince_sv}
  \end{subfigure}
  \caption{(a)-(c) Are the TSNE  of the full recorded frames set produced by the models trained with the loss specified in the sub caption using synthetic data. (d)-(f) Are the TSNE  of the full recorded frames set produced by the models trained with the loss specified in the sub caption with no synthetic data.}
  \label{fig:tsne_train}
\end{figure*}

\section{Data Cleaning}

\begin{algorithm}[bt]
    \caption{Motion Detection Algorithm}
    \label{algo:motion}
    \begin{algorithmic}
        \Require Frame list $F=[F_{1}\, ...\ F_{N}]$ of length $N$, gaussian blur kernel size tuple $k_G$, dilation kernel size tuple $k_D$, contour threshold $T_C$ and binarization threshold $T_B$
        \Function{MotionDet}{$F[\;]$}
          \State {$b$ $\gets$ $\varnothing$}
          \State {$a$ $\gets$ $\varnothing$}
          \For{$i \gets 1$ to $N$}
            \If{$b = \varnothing$}
                \State {$b$ $\gets$ {$F_{i}$}}
            \Else
                \State {$a$ $\gets$ {$F_{i}$}}
                \State {$a_{diff}$ $\gets$ $\textsc{AbsDiff}(a, b, k_G)$}
                \State {$c$ $\gets$ $\textsc{FindContours}(a_{diff}, k_D, T_B)$}
                \State {$s$ $\gets$ $\textsc{CalcAreas}(a, b, c)$}
                \State {$b,a$ $\gets$ $\textsc{Save}(a, b, s, T_C)$} 
            \EndIf
          \EndFor
        \EndFunction
    \end{algorithmic}
\end{algorithm}

Computer workflows and tutorials contain many repeated frames as well as frames with no relevant information, other than slight mouse movements. Although we can extract the frames at a low FPS rate, it is still not enough to remove the unwanted and unnecessary frames. Thus, we used a motion detection algorithm to filter out frames with small changes. This algorithm is a slight modification to the one in~\cite{motion}. \Cref{algo:motion} shows a pseudo code description of our algorithm \textsc{MotionDet}. 

\textsc{MotionDet} takes as input a frame list $F$ of length $N > 1$. First, it extracts from $F$ two consecutive frames $a$ and $b$. \textsc{AbsDiff} pre-processes the images using a Gaussian smoothing filter with kernel $k_G$. Then, it calculates the pixel-wise absolute difference between the pre-processed images to produce the image $a_{diff}$. The function \textsc{FindContours} takes $a_{diff}$, applies a dilation kernel $k_D$ to accentuate possible Regions Of Interest (ROIs) and binarizes the differences using some chosen threshold $T_B$ to produce a binary image. Later on, it follows Suzuki \etal's~\cite{contours} method to find ROIs $c$ on the binarized image. Dilation is a method that accentuates bright spots in images, by replacing the value of a pixel with the largest value in its neighborhood defined by the kernel $k_D$. In contrast to~\cite{motion} we dilate before binarization to allow for more sensitivity in the algorithm. Then, the function \textsc{CalcAreas} calculates the areas $s$ of the ROIs in $c$. 
Finally, \textsc{Save} writes the transition to memory if one of the ROIs' areas in $s$ is bigger than a threshold $T_C$. If true, $a$ and $b$ are saved and $a$ takes $\varnothing$ value while $b$ takes the value of $a$. Otherwise, $b$ stays the same. The process is repeated until the end of $F$ is reached.

The hyperparameter $T_C$ was chosen on a per-video-basis since smaller changes are considered of importance depending on the different situations, such as copying and pasting small text from one application to another. The other parameters values were fixed for all videos. $k_G$, $k_D$ have a size of $(5,5)$ and the threshold $T_B$ was set to $40$.

\end{document}